\documentclass[doublespace,11pt]{article}
\usepackage{graphicx}
\usepackage{setspace}
\usepackage{mathptm}
\usepackage{amsmath, amsthm}
\usepackage{subfigure}
\usepackage{epsfig}
\usepackage{my-apa-uiuc}

\begin{document}
\sloppy
\begin{spacing}{1}

\begin{titlepage}
\hspace{0.08in}
\begin{minipage}{\textwidth}
\begin{center}
\vspace*{3cm}
\begin{tabular}{c c c}
\hline
 & & \\
 & {\Huge The University of Algarve} & \\
 & & \\
 & {\Huge Informatics Laboratory} & \\
 & & \\
\hline
\end{tabular}\\
\vspace*{2cm}
{\Large
UALG-ILAB\\
Technical Report No. 200602 \\
February, 2006\\
}
\vspace*{3cm}
{\bf Revisiting Evolutionary Algorithms with On-the-Fly\\ 
Population Size Adjustment}\\
\addvspace{0.5in}
{\bf Fernando G. Lobo}, and {\bf Cl\'{a}udio F. Lima}\\
\vspace*{-0.1in}
\vspace*{4cm}
Department of Electronics and Informatics Engineering\\
Faculty of Science and Technology \\
University of Algarve \\
Campus de Gambelas\\
8000-117 Faro, Portugal\\
URL: {\verb http://www.ilab.ualg.pt }\\
Phone: (+351) 289-800900\\
Fax: (+351) +351 289 800 002 \\
\end{center}
\end{minipage}
\end{titlepage}

\title{\bf Revisiting Evolutionary Algorithms with On-the-Fly Population Size Adjustment}
\author{    {\bf Fernando G. Lobo}
\footnote{Also a member of IMAR - Centro de Modela\c{c}\~{a}o Ecol\'{o}gica.}\\
            \small UAlg Informatics Lab\\
            \small DEEI-FCT, University of Algarve\\
            \small Campus de Gambelas\\
            \small 8000-117 Faro, Portugal\\
            \small flobo@ualg.pt
\and
            {\bf Cl\'{a}udio F. Lima}\\
            \small UAlg Informatics Lab\\
            \small DEEI-FCT, University of Algarve\\
            \small Campus de Gambelas\\
            \small 8000-117 Faro, Portugal\\
            \small clima@ualg.pt
}
\date{}
\maketitle

\begin{abstract}
In an evolutionary algorithm, the population has a very important role
as its size has direct implications regarding solution quality, speed,
and reliability. Theoretical studies have been done in the past
to investigate the role of population sizing in evolutionary algorithms.
In addition to those studies, several self-adjusting population sizing
mechanisms have been proposed in the literature.
This paper revisits the latter topic and pays special attention to
the \textit{genetic algorithm with
adaptive population size} (APGA), for which several researchers have
claimed to be very effective at autonomously (re)sizing the population.

As opposed to those previous claims, this paper suggests a complete
opposite view. Specifically, it shows that APGA is not capable
of adapting the population size at all. This claim is supported on
theoretical grounds and confirmed by computer simulations.
\end{abstract}

\section{Introduction}
Evolutionary algorithms (EAs) usually have a number of control parameters
that have to be specified in advance before starting the algorithm itself.
One of those parameters is the population size, which in traditional EAs
is generally set to a specified value by the user at the beginning of the
search and remains constant through the entire run. Having to specify this
initial parameter value is problematic in many ways. If it is too small
the EA may not be able to reach high quality solutions. If it is too large
the EA spends too much computational resources.
Unfortunately, finding an adequate population size is a difficult task.
It has been shown, both theoretically and empirically, that the optimal
size is something that differs from problem to problem.
Moreover, some researchers have observed that at different stages of a
single run, different population sizes might be optimal.

Based on these observations, researchers have suggested various
schemes that try to learn a good population size during
the EA run itself~\cite{Lobo:05b}.
In a recent study~\cite{Eiben:04}, several adaptive population size methods
were compared head to head on a set of instances of the multimodal problem
generator~\cite{Spears:02}.
The winner of that competition was found to be the genetic algorithm with
adaptive population size~(APGA)~\cite{Back:00}, where the parameter-less
genetic algorithm~\cite{Harik:99} had the worst performance out of 5 contestant
algorithms, which included a simple GA with a fixed population size of 100.

This paper revisits the comparison between APGA and the parameter-less GA in
what is claimed to be a more fair basis than the one used before.
More important, the paper shows that APGA is not capable
of adapting the population size, a claim that is supported by theoretical and
empirical evidence.

The paper is structured as follows. The next section reviews two adaptive
population size methods based on age and lifetime.
Then, Section~\ref{apgaTheory} analyzes in detail how one of these algorithms,
the APGA, resizes the population through time.
In Section~\ref{theoryVerification} the analysis is
verified with experimental results. The parameter-less GA is described in
Section~\ref{parameterlessGA}. Section~\ref{criticalNote} makes a critical
note regarding a past comparative study of population (re)sizing methods,
and Section~\ref{theBigQuestion}
performs a comparison between APGA and the parameter-less GA for a class of
problems that has well-known population size requirements.
The paper finalizes with a summary and conclusions.

\section{Adaptive schemes based on age and lifetime} \label{review}

This section reviews two techniques for adapting the population size based
on the concept of age and lifetime of an individual. The first method
was proposed for adapting the population size of a generational GA,
while the second method is an extension of the first to allow
adaptive population sizing in a steady-state GA, incorporating also
elitism.

\subsection{GAVaPS}
The {\em Genetic Algorithm with Varying Population Size} (GAVaPS) was
proposed by \citeN{Arabas:94}. The
algorithm relies on the concept of {\em age} and {\em lifetime} of
an individual to change the population size from generation to
generation. When an individual is created, either during the initial
generation or through a variation operator, it has age zero.
Then, for each generation that the individual stays alive its age
is incremented by 1.

At birth, every individual is assigned a lifetime which corresponds
to the number of generations that the individual stays alive in the population.
When the age exceeds the individual's lifetime, the individual dies and is
eliminated from the population.
Different strategies to allocate lifetime to individuals can be used but the
key idea is to allow high-quality individuals to remain in the population
for longer number of generations than poor-quality individuals.
The authors suggested three different strategies: proportional, linear,
and bi-linear allocation. All those strategies relied on two parameters,
{\em MinLT} and {\em MaxLT}, which correspond to the minimum and maximum
lifetime value allowable for an individual.

At every generation, a fraction $\rho$ (called the {\em
reproduction ratio}) of the current population is allowed to reproduce.
Every individual of the population has an equal probability of being
chosen for reproduction. Thus, GAVaPS does not have an explicit selection
operator as traditional GAs do. Instead, selection is achieved indirectly
through the lifetime that is assigned to individuals. Those with
above-average fitness have higher lifetimes than those with below-average
fitness. The idea is that the better an individual is, the more time it
should be allowed to stay in the population, and therefore increase the
chance to propagate its traits to future individuals.
Figure~\ref{fig:gavaps} details the pseudo-code of GAVaPS.

\begin{figure}[!t]
\centering
\begin{verbatim}
 procedure GAVaPS
 begin
   t = 0;
   initialize pop(t);
   evaluate pop(t);
   while not termination-condition do
   begin
      t = t+1;
      pop(t) = pop(t-1);
      increase age of pop(t) members by 1;
      cross and mutate pop(t);
      evaluate pop(t);
      remove from pop(t) all individuals
         with age greater than their lifetime;
   end
 end
\end{verbatim}
\caption{Pseudocode of the genetic algorithm with varying population
size (GAVaPS).} \label{fig:gavaps}
\end{figure}

The authors tested GAVaPS on four test functions, compared its
performance with that of a simple GA using a fixed population size,
and observed that GAVaPS seemed to incorporate a self-tuning process
of the population size.
GAVaPS requires the user to specify the
initial population size, but the authors refer in their work that
GAVaPS is robust with respect to that, i.e., the initial population
size seemed to have no influence on the performance on the test
functions chosen. The same thing did not hold with the reproduction
ratio $\rho$. Different values of $\rho$ yielded different
performance of the algorithm. For the {\em MinLT} and {\em MaxLT}
parameters, Arabas et al. used 1 and 7 respectively.

\subsection{APGA}

The {\em Genetic Algorithm with Adaptive Population Size} (APGA)
proposed by \citeN{Back:00} is a
slight variation of the GAVaPS algorithm. The difference between the
two is that APGA is a steady-state GA, the best individual in the
population does not get older, and in addition to the selection
pressure obtained indirectly by the lifetime mechanism, APGA also
uses an explicit selection operator for choosing individuals to
reproduce. Thus, APGA uses a stronger selection pressure than
GAVaPS. An algorithmic description of APGA is presented in
Figure~\ref{fig:apga}.

\begin{figure}[!t]
\centering
\begin{verbatim}
 procedure APGA
 begin
   t = 0;
   initialize pop(t);
   evaluate pop(t);
   compute RLT for all members of pop(t);
   while not termination-condition do
   begin
      t = t+1;
      pop(t) = pop(t-1);
      decrement RLT by 1 for all but
          the best member of pop(t);
      select 2 individuals from pop(t);
      cross and mutate the 2 individuals;
      evaluate the 2 individuals;
      insert the 2 offspring into pop(t);
      remove from pop(t) those members with RLT=0;
      compute RLT for the 2 new members of pop(t);
   end
 end
\end{verbatim}
\caption{Pseudocode of the genetic algorithm with adaptive
population size~(APGA). RLT stands for {\em remaining
lifetime}.} \label{fig:apga}
\end{figure}

As opposed to the authors of GAVaPS, \shortciteANP{Back:00} set
the values of {\em MinLT} and {\em MaxLT} to 1 and 11, because according to
them, initial runs with different values indicated that {\em MaxLT=11}
delivers good performance. APGA also needs the initial population size to
be specified (B\"ack et al. used 60 individuals in their experiments).

At every iteration of the
steady-state GA, all individuals (except the best one) grow older by 1
unit. Thus, it's quite likely that after {\em MaxLT} iterations, most of
the individuals from the original population will have died and the only
ones that remain in the population are either: (1) the individuals
generated during the last {\em MaxLT} iterations, or (2) the best
individual from the initial population
(recall that the best individual does not get older).
In other words, after {\em MaxLT} iterations the population size will be
of order $\mathcal{O}(MaxLT)$. This argument has been hinted before~\cite{Lobo:05b}
and its correctness is confirmed in the next sections.

By not decreasing the remaining lifetime of the best individual, APGA
ends up using elitism because the best individual found so far always
remains in the population.
In~\shortciteN{Back:00}, the evolution of the population sizes through time is not shown,
but in all the reported experiments, the average population size at the end of
the runs were in the range between 7.8 and 14.1, which
confirms our reasoning that the population size in APGA tends to be of the
same order of {\em MaxLT}.
We will also confirm this reasoning by performing experiments under
different settings to observe how the population size in APGA changes
over time.

Similarly to GAVaPS, APGA can also use different lifetime strategies.
\shortciteA{Back:00} used a bi-linear strategy similar to the one
proposed for GAVaPS.
The bi-linear strategy is a commonly used strategy. For completeness, the
formula is shown below.

\begin{displaymath}
\left\{ \begin{array}{ll}
 MinLT + \eta \frac{fit[i]-MinFit}{AvgFit-MinFit} &
 \mbox{if $AvgFit \geq fit[i]$} \\
\\
 \frac{MinLT+MaxLT}{2} + \eta \frac{fit[i]-AvgFit}{MaxFit-AvgFit} &
 \mbox{if $AvgFit < fit[i]$}
        \end{array}
\right.
\end{displaymath}

\noindent
where $\eta=(MaxLT-MinLT)/2$.

\section{How APGA really works?} \label{apgaTheory}

Let's analyze in detail the population resizing mechanism of APGA.
Let $P(t)$ be the size of the population at generation $t$. Note that
in the pseudocode shown in Figure~\ref{fig:apga}, $P(t)$
refers to the size of the population at the end of the while loop.
At every generation 2 new individuals are created. Thus, the size
of the population at generation $t$ is given by the following
recurrence relation:
\begin{equation}
\label{eq:eq-popsize1}
      P(t) = P(t-1) + 2 - D(t) 
\end{equation}

\noindent
where $D(t)$ is the number of individuals which die at generation $t$.

Starting from an initial population size $P(0)$, which is a parameter
of the algorithm, it is possible to iterate the recurrence relation
and obtain the following expression for the population size at generation t:

\begin{equation}
\label{eq:eq-popsize2}
     P(t) = P(0) + 2~t - \sum_{i=1}^{t} D(i)
\end{equation}

\noindent
The summation $\sum_{i=1}^{t} D(i)$ denotes the number of individuals
who die in the first $t$ generations. Based on this observation, it is easy
to prove that regardless of the initial population size $P(0)$, the
population size after $MaxLT$ generations cannot be greater
than  $2~MaxLT + 1$.

\newtheorem{theorem}{Theorem}
\begin{theorem} Regardless of $P(0)$,
\label{theorem1}
\begin{displaymath}
P(MaxLT) \leq 2~MaxLT + 1
\end{displaymath}
\end{theorem}

\begin{proof}
With an exception for the best individual, $MaxLT$ is an upper bound on
the number of generations than any given individual is allowed to live.
Thus, after $MaxLT$ generations we can be sure that all
but the best member from the initial population will be dead. That is,

\begin{displaymath}
\sum_{i=1}^{MaxLT} D(i) \geq P(0) - 1
\end{displaymath}

\noindent
Using this result together with equation~\ref{eq:eq-popsize2} yields,

\begin{eqnarray*}
  P(MaxLT) & = & P(0) + 2~MaxLT - \sum_{i=1}^{MaxLT} D(i)\\
           & \leq & 2~MaxLT + 1
\end{eqnarray*}
which concludes the proof.
\end{proof}

Now let us prove that $2~MaxLT + 1$ remains an upper bound for
the population size for the remaining generations.

\begin{theorem}
For all $t \geq MaxLT$,
\begin{displaymath}
P(t) \leq 2~MaxLT + 1
\end{displaymath}
\end{theorem}

\begin{proof}
One just needs to notice that the exact same reasoning used to prove
Theorem~\ref{theorem1} can be used to prove the same thing
assuming that the starting point is not the initial generation, but
instead, some arbitrary generation $k$. That is, regardless of $P(k)$,

\begin{displaymath}
P(k+MaxLT) \leq 2~MaxLT + 1
\end{displaymath}

Thus, using induction with $k=0$ as a base case, we prove that the
upper bound $2~MaxLT + 1$ holds for all $k>0$.
\end{proof}

\begin{figure}[t]
  \center
  \epsfig{file=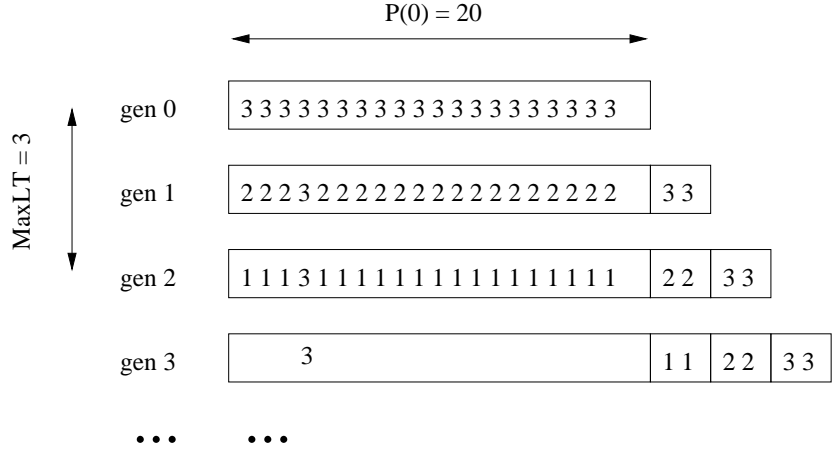,width=0.7\columnwidth}
  \caption{An example with $MaxLT=3$ of the evolution of the population
size in APGA. The numbers in the figure denote an upper bound for the
remaining lifetime of the individuals.}
  \label{fig:fig-proof}
\end{figure}

The proofs that we have seen are relatively straightforward. Nevertheless,
in order to make them more easily understandable,
Figure~\ref{fig:fig-proof} depicts schematically an example (with
$P(0)=20$ and $MaxLT=3$) of what might be the state of the population
after $MaxLT$ generations have elapsed. The example shown corresponds to
the upper bound for the size of the population at generation $maxLT$,
which is obtained when all individuals are assigned the maximum
lifetime value during their creation. The numbers in the figure denote
the remaining lifetime (RLT) of the individuals.
When an individual is created it is assigned a RLT of 3. Then, at every
generation, 2 new individuals are created (and also assigned an RLT of 3)
and the remaining ones have their RLT value decremented by 1. An exception
occurs for the best individual of the previous generation. In the example,
we are assuming the fourth individual (from left to right) to be the best one.
Notice that when going from generation to generation, the new best individual
can only be the previous best or one of two newly created solutions.
The upper bound of $2~MaxLT + 1$ corresponds to the situation where the best
solution remains the same for the last MaxLT generations, as depicted in the
figure.

In summary, after $MaxLT$ generations, the population size in APGA
is upper bounded by $2~MaxLT + 1$, and from that point on until the
end of the search, the population size will not raise beyond that
bound.

Notice that what we have shown is an upper bound on the maximum
population size. To discover that upper bound we had to be conservative
and assume that all individuals that are created are able to stay in the
population for $MaxLT$ generations.
In practice, what is likely to occur in a real APGA simulation is
that the actual population size will be somewhat less than that.
Due to the effects of selection, it is not clear what is the expected
number of generations that an individual stays in the population, but
as a very crude approximation, we could say that it should be a value
close to $AvgLT = (MinLT+MaxLT)/2$.
This thought experiment, suggests that if we replace $MaxLT$ by $AvgLT$
in the upper bound expression, we can get an approximation of the
steady state population size of APGA.

\newtheorem{conj}{Conjecture}
\begin{conj} For $t \geq MaxLT$, the size of the population is
approximately $MinLT + MaxLT + 1$.
\end{conj}

\begin{eqnarray*}
     P(t) & \approx & 2~AvgLT + 1\\
          &       = & 2~(MinLT+MaxLT)/2) + 1\\
          &       = & MinLT + MaxLT + 1
\end{eqnarray*}

At this point, it is time to do some simulations to confirm the theory.

\section{Verifying the theory} \label{theoryVerification}

APGA was tested on 3 problem instances (1, 50, and 100
peaks) of the multimodal problem generator used by~\shortciteN{Eiben:04}.
This generator
creates problem instances with a controllable number of peaks (the
degree of multi-modality). For a problem with $\cal P$ peaks, $\cal
P$ $L$-bit strings are randomly generated. Each of these strings is
a peak (a local optima) in the landscape. Different heights can be
assigned to different peaks based on various schemes (equal height,
linear, and so on).
To evaluate an arbitrary individual $\bar x$, first locate
the nearest peak in Hamming space, call it $Peak_{n}(\bar x)$.
%
%
Then the fitness of $\bar x$ is the number
of bits the string has in common with that nearest peak, divided by $L$,
and scaled by the height of the nearest peak.

\begin{displaymath}
f(\bar x) = \frac{L-Hamming(\bar x,Peak_{n}(\bar x))}{L}
            \cdot Height(Peak_{n}(\bar x))
\end{displaymath}

\begin{figure}[t]
  \center
  \epsfig{file=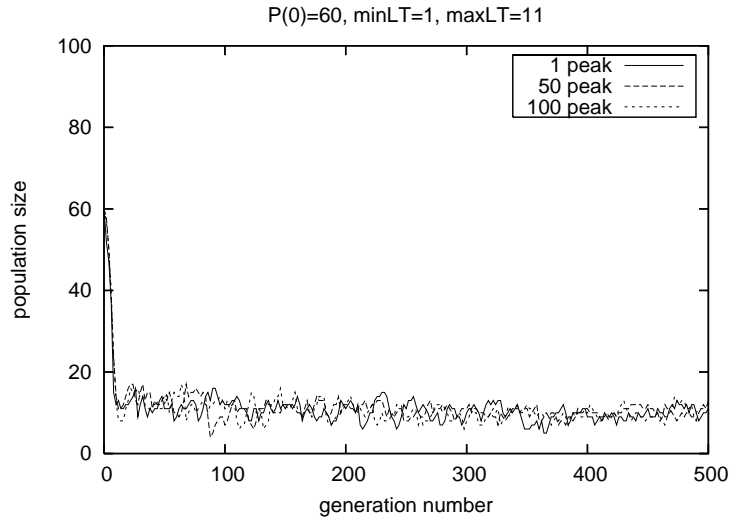, width=0.6\columnwidth}
  \caption{Starting from an initial population size of 60, after $MaxLT=11$
generations the population size is fluctuating around 11 and never raises
past $2~MaxLT+1=23$.}
  \label{fig:n60-lt1-11}
\end{figure}

\begin{figure}[t]
  \center
  \epsfig{file=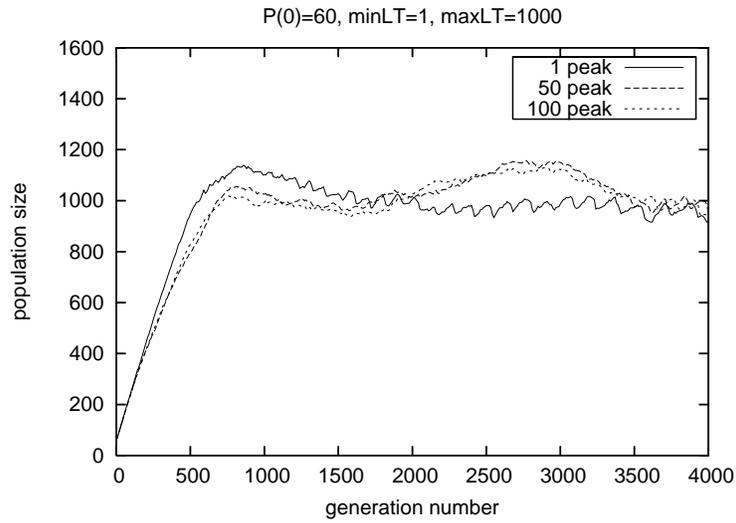, width=0.6\columnwidth}
  \caption{Using $MaxLT=1000$, the population raises initially from 60
           to 1000 and then stabilizes around that value.}
  \label{fig:n60-lt1-1000}
\end{figure}

\begin{figure}[t]
  \center
  \epsfig{file=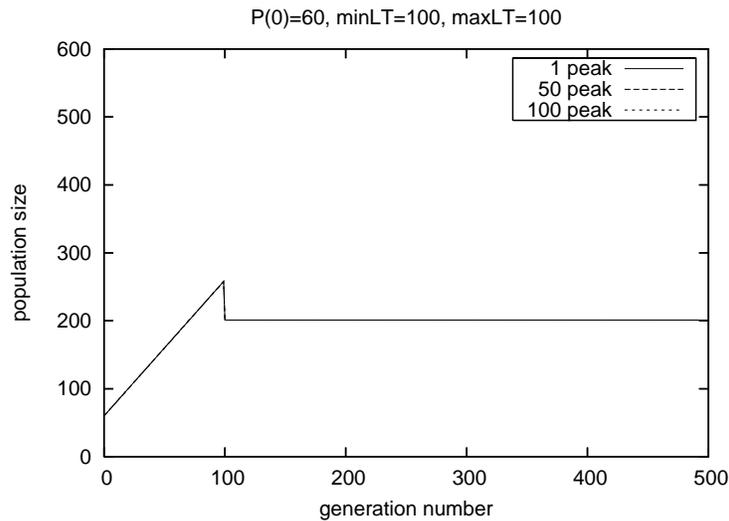, width=0.6\columnwidth}
  \caption{This example comes very close to the upper bound. Using
$MinLT=MaxLT=100$, the population keeps raising by 2 individuals for the
first 99 generations. At generation 100, drops to a value of 200 (or 201).}
  \label{fig:n60-lt100-100}
\end{figure}

Figures~\ref{fig:n60-lt1-11},
\ref{fig:n60-lt1-1000}, \ref{fig:n60-lt100-100}, monitor the
population size as time goes by using an initial population size of
60, and minimum and maximum lifetime values of 1 and 11 (like
suggested by their authors~\cite{Back:00,Eiben:04}, 1 and 1000, and
also 100 and 100. The latter setting was tested deliberately to
verify the upper bound for the size of the population. Note that
when $MinLT=MaxLT$, all individuals are assign that same value as
their lifetime, and that should correspond to the situation where
the population size stays as close as possible to the upper bound of
$2~MaxLT + 1$, as depicted schematically in
figure~\ref{fig:fig-proof}.

For completeness, we also use the exact same settings for the other 
parameters and
operators as those used by \shortciteA{Eiben:04}: two-point crossover with
$P_{c}=0.9$, bit-flip mutation with $P_{m}=1/L=0.01$, and binary
tournament selection.

\begin{figure}[t]
  \center
  \epsfig{file=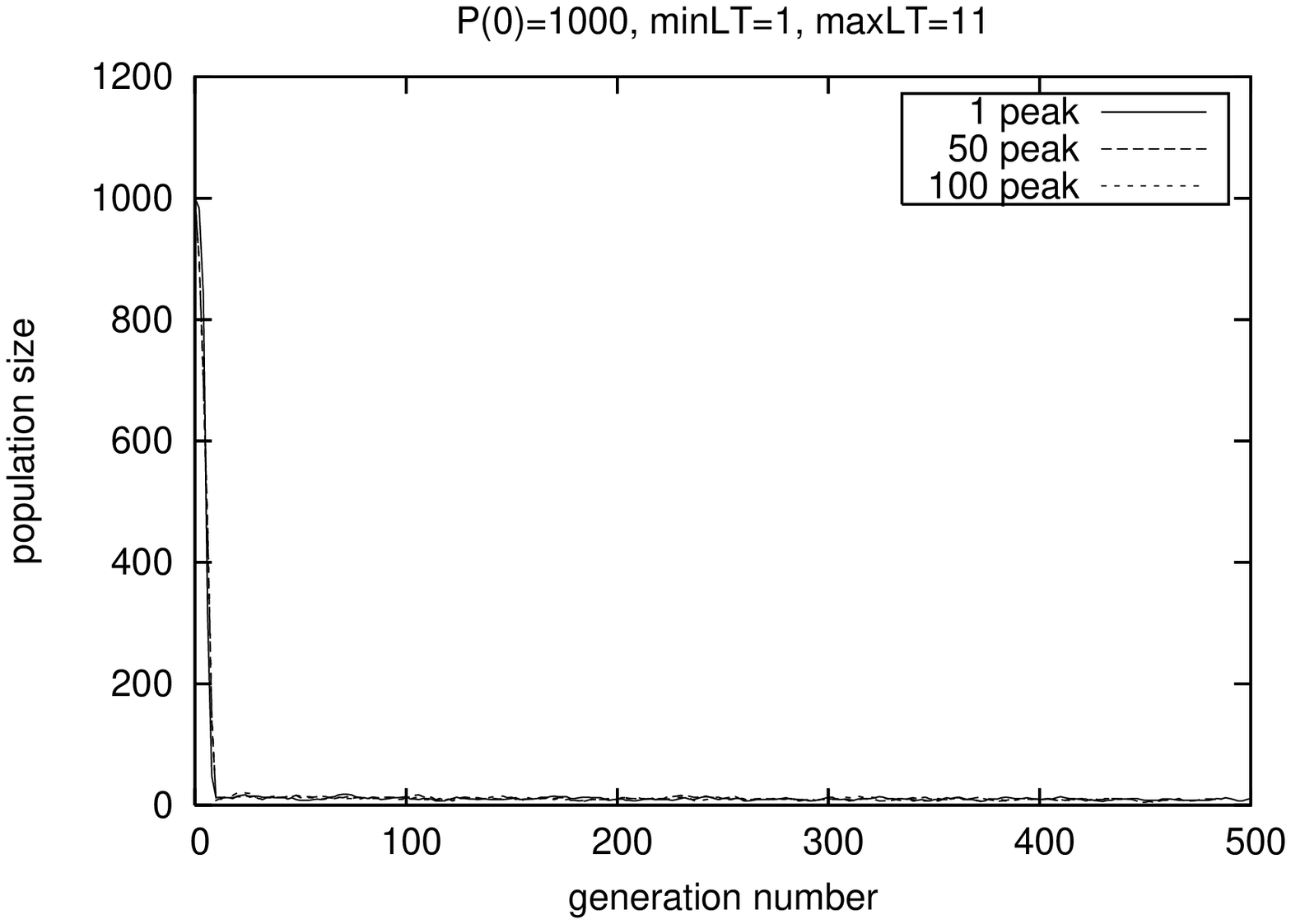, width=0.6\columnwidth}
  \caption{P(0)=1000, MinLT=1, MaxLT=11.
           Same behavior as in figure~\ref{fig:n60-lt1-11}.}
  \label{fig:n1000-lt1-11}
\end{figure}

\begin{figure}[t]
  \center
  \epsfig{file=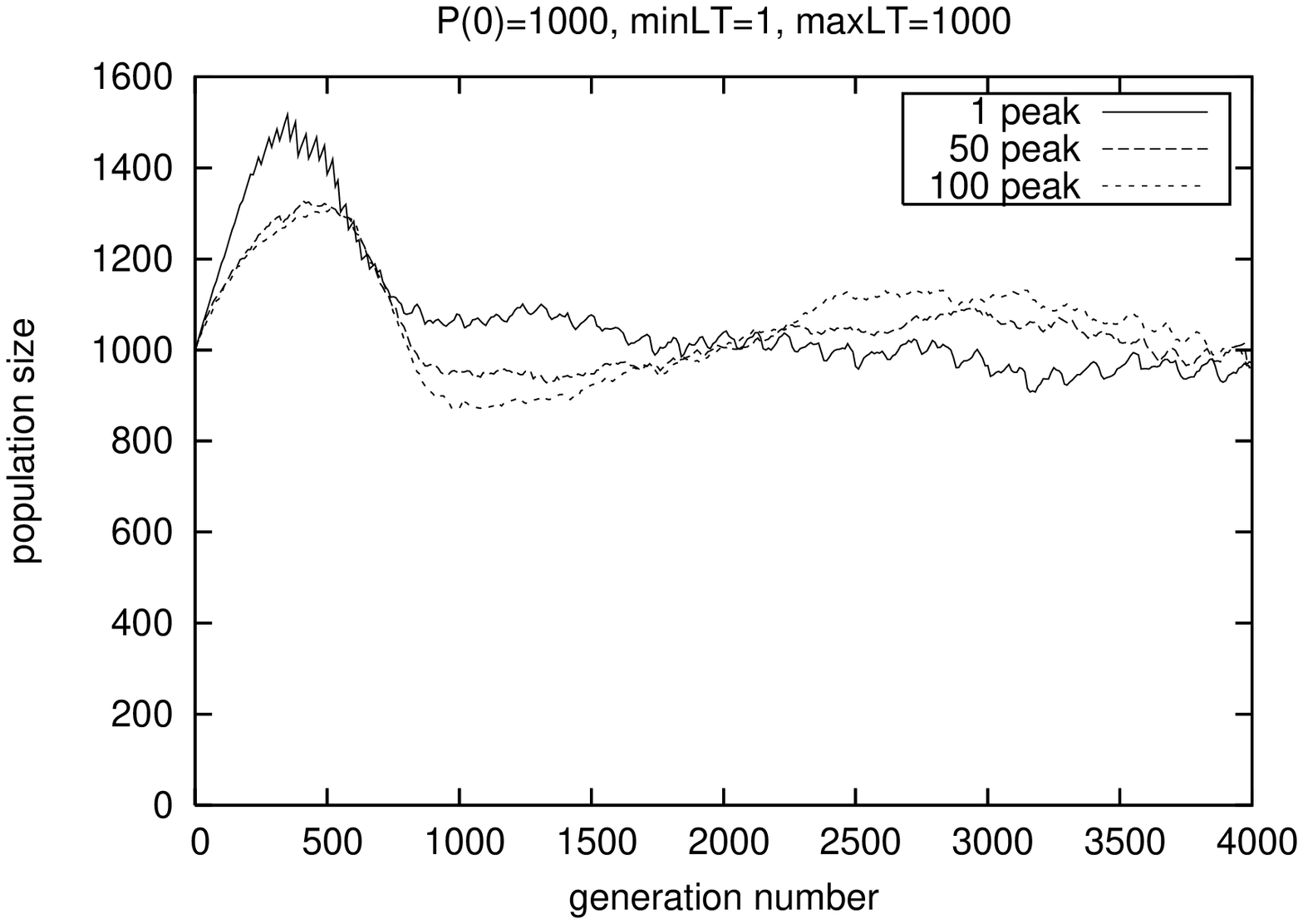, width=0.6\columnwidth}
  \caption{P(0)=1000, MinLT=1, MaxLT=1000.
           Same behavior as in figure~\ref{fig:n60-lt1-1000}.}
  \label{fig:n1000-lt1-1000}
\end{figure}

\begin{figure}[t]
  \center
  \epsfig{file=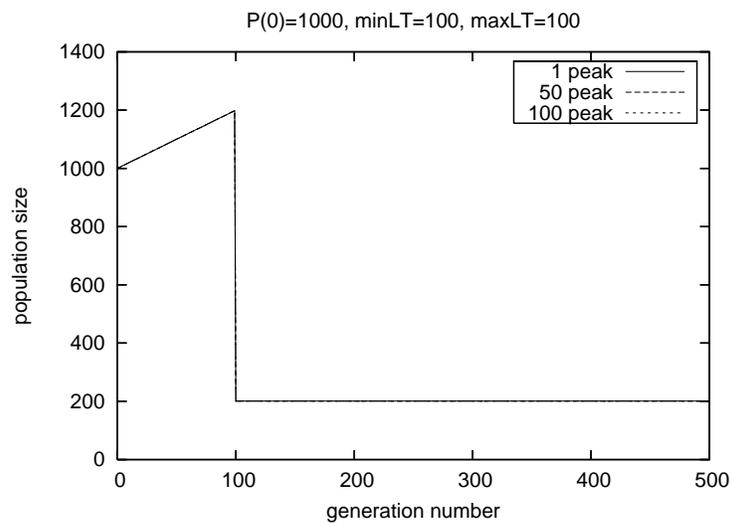, width=0.6\columnwidth}
  \caption{P(0)=1000, MinLT=100, MaxLT=100.
           Same behavior as in figure~\ref{fig:n60-lt100-100}.}
  \label{fig:n1000-lt100-100}
\end{figure}

We have also performed similar experiments starting with an initial
population size value of 1000 individuals (see
figures~\ref{fig:n1000-lt1-11}, \ref{fig:n1000-lt1-1000},
\ref{fig:n1000-lt100-100}). Again, the theory is confirmed.
Further
experiments were also done with the Java implementation provided by
\shortciteN{Eiben:04} at \texttt{http://www.cs.vu.nl/ci}.
These latter experiments were done to make sure that no detail was
missing from an eventual lack of understanding on our part regarding
the mechanics of the algorithm. The results were identical to the
ones obtained with our own implementation.

These results constitute a strong evidence that APGA is not capable
of adapting the population size. Independently of the problem being solved,
after $MaxLT$ generations have elapsed ($2~MaxLT$ function evaluations),
the population size tends to fluctuate around a value close
to $MinLT+MaxLT$.
In other words, the parameters $MinLT$ and $MaxLT$ end up acting
as a camouflage of the population size.

\clearpage

We now leave APGA for a little while and briefly review another algorithm
that does not require the specification of a fixed population size,
the parameter-less GA. Later in the paper, both algorithms will tested
head-to-head.

\section{Parameter-less GA} \label{parameterlessGA}

The parameter-less GA introduced by \citeN{Harik:99}
was developed with the assumption that solution quality
grows monotonously with the population size.
That is, if after some
number of generations $t$, a GA is able to reach a solution quality
$q$ with some population size $N_{1}$, then (in a statistical sense)
it would also be able to reach that same solution had it started
with a population size $N_{2} > N_{1}$.
Based on that observation, \citeANP{Harik:99} suggested a scheme that
simulates an unbounded (potentially infinite) number of populations
running in ``parallel'' with exponentially increasing sizes.
Their scheme gives preference to smaller sized populations by allowing
them do do more function evaluations than the larger populations.
The rationale is that all other things being equal, a smaller sized
population should be preferred.
After all, if a GA can reach a good
solution quality (goodness depends on the stopping criteria, of course)
with a population size $N_{1}$, then why bother spending more efforts with
a population size $N_{2} > N_{1}$.

Initially the algorithm only has one population whose size is a very
small number $N_{0}$ (say 4).
As time goes by, new populations are spawned and some can be deleted
(more about that later). Thus, at any given point in time, the algorithm
maintains a collection of populations.
The size of each new population is twice as large as the previous
last size. The parameter-less GA does have one parameter!
(although \citeANP{Harik:99} fixed its value to 4).
That parameter (call it $m$) tells how much preference the algorithm
gives to smaller populations. Specifically, it tells how many generations
are done by a population before the next immediate larger population has
a chance to do a single generation.

\begin{figure}
\centering
  {\epsfig{file=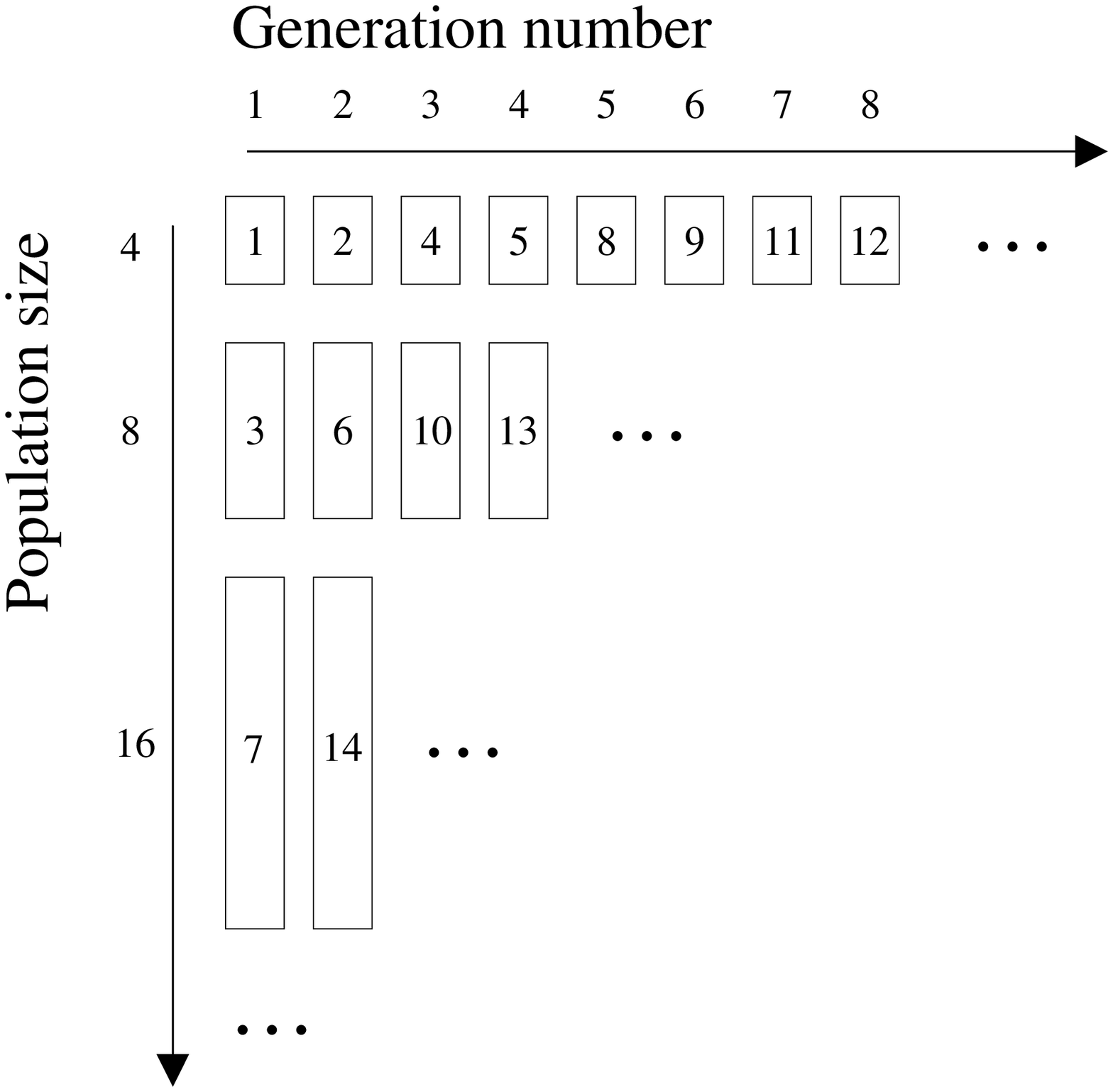,width=0.6\linewidth}}
\caption{The parameter-less GA simulates an unbounded number of populations.}
\label{fig:plessGA-mechanics}
\end{figure}

An example is helpful to illustrate the mechanism.
Figure~\ref{fig:plessGA-mechanics} depicts an example with $m=2$.
Notice that the number of populations is unbounded. The figure shows
the state of the parameter-less GA after 14 iterations. At that point,
the algorithm maintains three populations with sizes 4, 8, and 16. Notice
how a population of a given size does $m=2$ more generations
than the next larger population. In the figure, the numbers inside the
rectangles denote the sequence in which the generations are executed
by the algorithm. The next step of the algorithm (not shown in the figure)
would spawn a new population with size 32.

This special sequence can be implemented with a $m$-ary counter as
suggested by \citeN{Harik:99}, and also with a
somewhat simpler implementation as suggested by \citeN{Pelikan:04a}.

In addition to maintaining an unbounded collection of populations, the
algorithm uses a heuristic to eliminate populations when certain events
occur.
In particular, when the average fitness of a population is greater
than the average fitness of a smaller sized population, the algorithm
eliminates the smaller sized one. The rationale for taking this decision
is based on the observation that the larger population appears to be
performing better than the smaller one, and it is doing so with less
computational resources (recall that the algorithm gives preference to
smaller populations). Thus, whenever such an event occurs, \citeANP{Harik:99}
reasoned that that was a strong evidence that the size of the smaller
population was not large enough, and the algorithm should not waste any
more time with it.
By doing so, the algorithm maintains an invariant that
the average fitness of the populations are in decreasing order, with
smaller sized populations having higher average fitness than larger
populations.

In the absence of mutation, the parameter-less GA also eliminates
populations that converge (convergence meaning that the whole population
consists of copies of identical individuals) since its not possible to
generate new individuals thereafter.
Elsewhere it has been shown that the worst case time complexity of the
parameter-less GA is only within a logarithmic factor with respect with
a GA that starts with an optimal population size~\cite{Pelikan:99c}.

\section{A critical note on experimental research methodology} \label{criticalNote}

A comparative study of EAs with on-the-fly population size
adjustment has been made in a recent paper~\cite{Eiben:04}. In that
study, in addition to APGA and the parameter-less GA, three
other algorithms entered the competition: a traditional GA with a
fixed population size of 100 (TGA), the GA with random variation of
population size (RVPS), and a newly introduced algorithm called
PRoFIGA. A description of the algorithms is available
elsewhere~\cite{Eiben:04,Lobo:05b}.

To compare the algorithms, the multimodal random peak problem
generator from \citeN{Spears:02} was used.
%
%
%
%
%
\shortciteANP{Eiben:04} compared the performance of the 5 algorithms on problems
with different number of peaks ranging from 1 to 1000. For each problem
instance, 100 independent runs were done and the following 3 performance
measures were recorded:

\begin{itemize}
\item Success Rate (SR), the percentage of runs in which the
global optimum was found.
\item Mean Best Fitness (MBF), the average of the best fitness in the last
population over all runs.
\item Average number of Evaluations to a Solution (AES), the number of
evaluations it takes on average for the successful runs to find the
optimum. If a GA has no success (SR=0) then the AES measure is undefined.
\end{itemize}

With the exception of population size, all algorithms used the same
parameter settings and operators: two-point crossover with $P_{c} = 0.9$,
bit-flip mutation with $P_{m} = 1/L$, binary tournament selection, and
delete worst-2 replacement strategy. For all algorithms the GA model
was steady-state, not generational.

The experiments were performed on 100-bit string problems and the
contestants were allowed to run until either they found the global
optimal solution, or a maximum of 10000 function evaluations elapsed.
The reader is referred to the original source~\cite{Eiben:04,Valko:03} for more details.

The authors run the experiments and claimed the superiority of APGA,
followed closely by PRoFIGA. It is our strong belief that such conclusions
are abusive and can be turned upside down very easily. A number
of issues should be highlighted.

\begin{enumerate}
\item Only allowing the algorithms to run for 10000 function evaluations,
is not sufficient to draw any conclusion as to what might happen if
the algorithms are allowed to run for longer (or shorter) time
spans.
The very low success rate obtained for the more difficult problems
suggest that the 10000 function evaluations were not sufficient to
let the algorithms display their ability in adapting or not the
population size.

\item One of the contestants, PRoFIGA, requires the specification of
7 additional parameters that were tuned a-priori for these same
problems~\cite{Valko:03}.

\item The parameter-less GA was not properly implemented because the
maximum population size was upper bounded, and the parameter-less GA
has no such bound. Quoting~\shortciteN{Eiben:04}, ``the parameter-less GA
is run in parallel with the following 8 population sizes: 2, 4, 8,
16, 32, 64, 128, 256.'' In addition to that, the authors should have
taken care of only incrementing the 4-base counter after doing
$N/2$ generations of the steady state GA ($N$ being the population size)
because that is the equivalent of one generation in a generational GA,
otherwise large populations are created very quickly violating the
principle that more fitness function evaluations are given first to
small sized populations.
\end{enumerate}

In addition to the above mentioned flaws, the class of problems
generated by the random peak problem generator is probably not the
most appropriate for assessing the performance of evolutionary
algorithms as shown elsewhere~\cite{Lobo:06b}.
In any case,
and for the purpose of demonstrating that the comparative study
presented by~\shortciteANP{Eiben:04} is unfair, let us redo the experiments
for 2 instances of the multimodal problem generator, one with 50 and
another with 100 peaks. For each instance, 100 independent runs are
performed. This time, however, instead of letting the algorithms run
until a maximum of 10 thousand function evaluations, we let them run
for a maximum of 1 million evaluations.

The APGA uses the exact same settings as those reported
by~\shortciteANP{Eiben:04}: MinLT=1, MaxLT=11, binary tournament selection,
2-point crossover with probability $P_{c}=0.9$, and bit-flip
mutation with $P_{m}=1/L=0.01$ The parameter-less GA uses the exact
same settings with the exception of the selection rate. It uses a
tournament size of 4. The reason why we do so is to make the two
algorithms have more or less the same selection pressure (note that
APGA has also an extra selection pressure incorporated in its
lifetime mechanism). Notice also that in the original parameter-less
GA, \citeANP{Harik:99} recommended a crossover
probability of $P_{c}=0.5$, but we are ignoring those
recommendations here in order to run both algorithms under similar
conditions. \citeANP{Harik:99} also did not give any recommendations
regarding mutation rates, but a small mutation rate of $P_{m}=1/L$
cannot possibly do much harm. The performance measures for the 50
and 100 peak problem instances are shown in
table~\ref{tab:peaks50-100}.

\begin{table}
\caption{APGA versus Parameter-less GA under a maximum of 1 million
function evaluations.}
\label{tab:peaks50-100}
\begin{center}
\begin{tabular}{|r|c|c|c|}
\hline
Problem & Measure & APGA      & Parameter-less GA\\
\hline \hline
          & SR  & 33\%        & 100\%     \\ \cline{2-4}
 50 peaks & AES & 1112      & 40142   \\ \cline{2-4}
          & MBF & 0.982     & 1.000   \\ \hline
          & SR  & 17\%        & 96\%      \\ \cline{2-4}
100 peaks & AES & 1282      & 74654   \\ \cline{2-4}
          & MBF & 0.976     & 0.999   \\
\hline
\end{tabular}
\end{center}
\end{table}

By allowing the algorithms to run for a longer time, the conclusions
are completely different from those obtained by~\citeANP{Eiben:04}. On
those occasions where APGA reaches the highest peak, it does so very
fast. The problem is that APGA is not consistent in reaching
the highest peak, not even with 1 million function evaluations. As
opposed to that, the parameter-less GA is capable of achieving high
success rates, but it can only do it if we give it enough time to do
so.

Let us observe now what are were the population sizes needed by both
algorithms to reach the highest peak. The population size needed by
APGA to reach the highest peak is on average 9.9 (for 50 peaks)
and 10.0 (for 100 peaks). That is expected because $MinLT=1$ and
$MaxLT=11$. For the parameter-less GA the average population size
is 182.5 (for 50 peaks) and 283.4 (for 100 peaks). The parameter-less
GA however exhibits a high variance. Sometimes solves the problem as
quickly as APGA using a population of size 8, and sometimes needs
around 800000 evaluations using a population of 2048.

The reason why this happens is because of the characteristics of the
multimodal problem generator. As explained
elsewhere~\cite{Lobo:06b}, instances with a large number of peaks
can only be solved reliably by an EA if a large population size is
used. Otherwise, it can only be solved due to luck, and in those
cases, it can be solved very fast, even with a very small population
size. The intuition behind this reasoning comes from the observation
that when an EA attempts to solve a problem with multiple peaks, it
fairly quickly concentrates the population around a single peak.
From that point on, the peak can be easily climbed, and obviously it
can be climbed faster if a small population size is used. The
problem though, is that it is unlikely for the EA to focus its
population on the best peak. This reasoning also explains the
observation of \shortciteANP{Eiben:04} that the AES measure for APGA did not
seem much affected by the number of peaks (recall that the AES
measure only averages the successful runs).

It should be pointed out that the reason why the parameter-less GA
achieved a high success rate in these experiments if not entirely due
to its capability of adapting the population size. What is more responsible
for the high success rate is the fact that the parameter-less GA maintains
a collection of populations, and in effect, each one can be seen as an
independent GA run with a different population size. By doing so,
the parameter-less GA increases its chances that one of those populations
ends up focusing around the basin of attraction of the best peak due
to pure luck.

We now look at another type of problem, an instance of an additively
decomposable problem, which has well known population sizing
requirements~\cite{Harik:99b,Goldberg:92}. We will be looking at how
APGA and the parameter-less GA try to solve the problem.

\section{To adapt or not to adapt} \label{theBigQuestion}

Although it can be argued that real world problems are unlikely to
be completely decomposable, this class of problems allow researchers
to exploit modularity, hierarchy, and bounded difficulty, in a
controllable manner. Moreover, this class of problems is the only one
for which theoretical population sizing models exist. Thus, they are
a natural candidate for testing self-adjusting population sizing
mechanisms.

An example of a decomposable problem is a function composed of
multiple deceptive sub-functions.
Deceptive functions normally have one or more deceptive optima that
are far away from the global optimum and which misleads the search
in the sense that the attraction area of the deceptive optima is
much greater than the one of the optimal solution.  A well known
deceptive function is the k-trap function \cite{Deb:93a} defined as
follows:

\begin{equation}
  f_{trap}(u) = \left\{ \begin{array}{ll}
      1                       & \textrm{if $u$ = $k$}\\
      1-d-u*\frac{1-d}{k-1}   & \textrm{otherwise}
    \end{array} \right.
\end{equation}


\noindent
where $u$ is the number of 1s in the string, $k$ is the size of the
trap function, and $d$ is the fitness signal between the global
optimum and the deceptive optimum.

If the whole problem is deceptive, there is no way to find the global
optimum efficiently by any algorithm because the problem is akin to
a needle in a haystack. But if the level of deception is bounded to
a few number of bits, the problem becomes solvable by GAs. A commonly
used bounded deceptive problem consists of a concatenation of $m$ copies
of a $k$-bit trap function. Then, the fitness of a solution is the sum
of the contributions of the $m$ trap functions.

\begin{equation}
  f(X) = \sum_{i=0}^{m-1} f_{trap}(x_{ki},x_{ki+1},\ldots,x_{ki+k-1}).
\end{equation}

On this type of problems, GAs are able to reliably find the global
optimum in an efficient way, provided that the population is
properly sized~\cite{Goldberg:92,Harik:99b}, and also assuming that
the crossover operator is not too disruptive. Since this problem has
well known population sizing requirements, it is a natural candidate
for testing the ability of EAs incorporating self-adjusting
population sizing mechanisms.

We will be testing both APGA and the parameter-less GA on a single instance
of the concatenated trap problem, an 80-bit problem
consisting of $m=20$ concatenated 4-bit trap functions (we use
$d=1/k=0.25$). The goal of the experiments is not to make a strict comparison
of the algorithms, nor saying that one is better than the other. Instead,
we want to illustrate how the population resizing mechanisms of the
algorithms behave on a problem that is known to have minimal population
sizing requirements in order to be solved efficiently.
For both algorithms we use the same parameter settings as described
in the previous section.

We do 100 independent simulations and stop the algorithms either when
the global optimum is found, or when a maximum of 1 million function
evaluations is reached. The APGA failed to solve a single run to
optimality. A reason for that? APGA is not capable of adapting the population
size and tries to solve the problem with populations sized around
9-10 individuals, a value of the same order of magnitude as $MaxLT$.
The parameter-less GA (we use $m=4$ as in suggested by~\citeANP{Harik:99}),
on the other hand, was able to learn that it
had to raise the population well beyond that, and obtained 100\% success
rate, taking on average 147 thousand function evaluations to reach
the global optimum. The population size needed by the parameter-less
GA to reach the optimum was on average 840, and the distribution was
256 (2/100 runs), 512 (43/100 runs), 1024 (50/100 runs), and 2048
(5/100 runs).

\begin{figure}[t]
  \center
  \epsfig{file=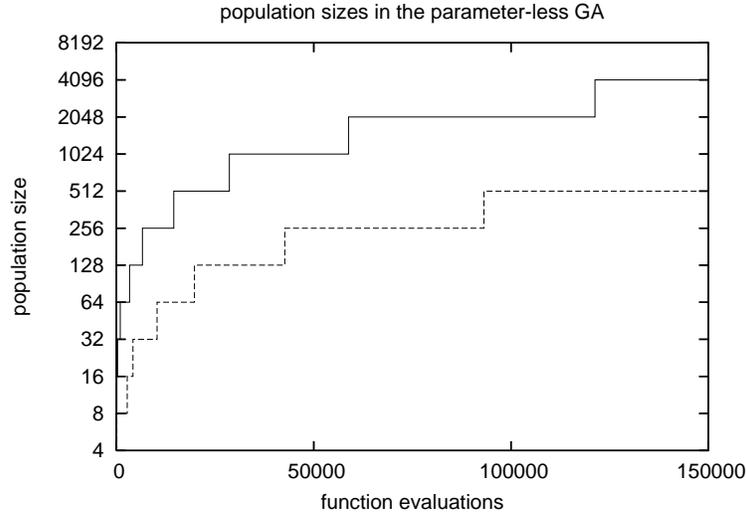,width=0.6\columnwidth}
  \caption{Range of population sizes maintained by the parameter-less GA.
The dashed line is the minimum population size, and the solid line is
the maximum population size maintained by the algorithm at any given point
in time.}
  \label{fig:fig-plessGA-popsizes}
\end{figure}

Figure~\ref{fig:fig-plessGA-popsizes} shows how the ranges of population
sizes maintained by the parameter-less GA evolves as time goes by.
The figure shows what happens on a single run alone. Other runs have
a similar behavior. In the figure there are two lines. The bottom one is
for the lowest sized population maintained by the algorithm. The line
above it is for the largest sized
population maintained by the algorithm at any given point in time.
Notice how early in the run, the algorithm is using small sized populations,
but fairly quickly it detects that those small sizes are not enough and
eliminates them. Each vertical step in the lower line corresponds to an
event where a population is deleted (because a larger population has a
higher average fitness). Regarding the top line, each vertical step
corresponds to an event where a new population is being created for the
first time. For example, the population of size 1024 was first created
when the algorithm had already spent around 30 thousand evaluations.

The lower line represents in some sense the current guess of the
parameter-less GA as to what might be a good population size for the problem
at hand. In this particular run, the algorithm reached the global optimum
with the population size of 1024, and did so after spending a total of 149
thousand function evaluations.

\begin{figure}[t]
  \center
  \epsfig{file=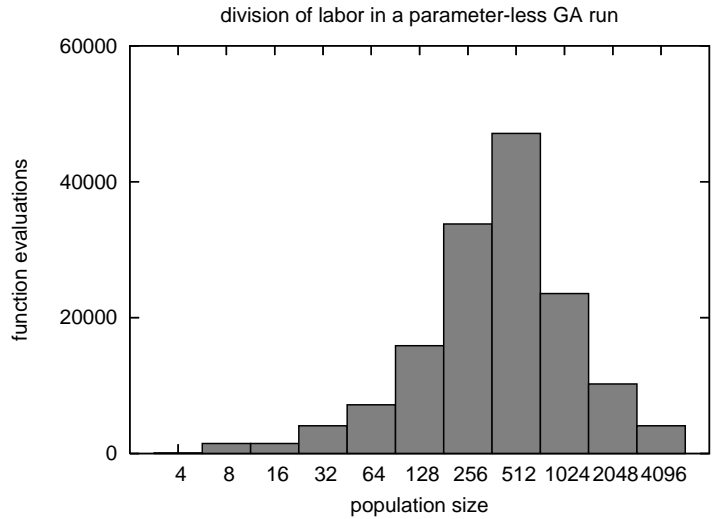,width=0.6\columnwidth}
  \caption{Number of function evaluations spent by each population size
during an entire run of the parameter-less GA.}
  \label{fig:division-of-labor}
\end{figure}

\begin{figure}[t]
  \center
  \epsfig{file=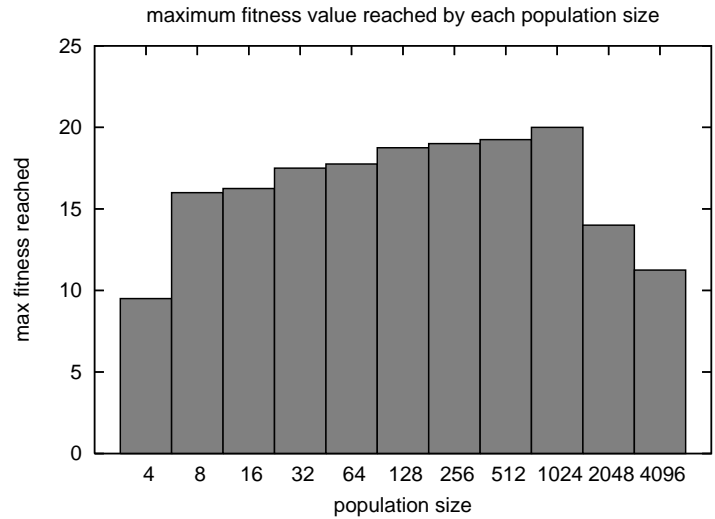,width=0.6\columnwidth}
  \caption{Maximum fitness value reached by each population size
during an entire run of the parameter-less GA.}
  \label{fig:maxfit-reached}
\end{figure}

For that same run of the parameter-less GA,
figures~\ref{fig:division-of-labor} and \ref{fig:maxfit-reached}
show the division of labor (how many function evaluations were spent by
each population size) as well as the maximum fitness value reached by
each population size through the entire run. In this particular run,
the population of size 1024 was able to reach the global optimum. All the
other evaluations are in some sense an overhead for the parameter-less
GA, a price that it had to pay in order to remove the guessing and tweaking
from the user. Notice also how larger populations are able to reach higher
fitness values. The exceptions occur with the larger populations (2048 and
4096 in this case) because they are still in their early stages of evolutions
and not many evaluations were spent with those population sizes yet.

\begin{table}
\centering
\caption{Performance measures. The APGA uses an initial population size
of 2000, $MinLT=1$, $MaxLT=2000$)}
\begin{tabular}{|c|c|c|} \hline
\label{tab:traps}
    & tweaked APGA      & Parameter-less GA\\ \hline \hline
SR  & 94                & 100    \\ \hline
AES & 35786             & 147285 \\ \hline
APS & 1685              & 840    \\
\hline\end{tabular}
\end{table}

Now let us give a little help to APGA by repeating the experiments
with $MaxLT=2000$. Presumably, this time APGA should be able to
solve the problem because it is going to ``adjust'' (if we can say
that) the population size to a value of that order of magnitude.
Contrary to our intuition, the APGA failed to solve the problem
reliably even with $MaxLT=2000$. A closer look at what was going on
revealed what was wrong. We were using an initial population size of
60 (recall that APGA also has an initial population size parameter).
Although the algorithm quickly raises those 60 individuals up to
2000, what happens is that by the time the population sets around
that value, it has already been affected by a substantial amount of
selection pressure, and some sub-structures end up having a low supply
of raw building blocks~\cite{Goldberg:01}.

The problem was fixed by setting the initial population size to
2000 (rather than 60) so that APGA can start right from the beginning
with a sufficient amount of raw building blocks.
As expected, the APGA with $P(0)=2000$, $MinLT=1$ and $MaxLT=2000$
solves the problem with a success rate of 94\% (still missed
6 runs), and its AES measure is 35786, faster than the parameter-less GA.
This time, however, the average population size used by APGA to solve
the problem was 1685. Again, a value of the same order of magnitude as
$MaxLT$.

What we are showing with these results is not that one algorithm is faster
than the other.
What our results do show is that the parameter-less GA is
capable or learning a good population size for solving the problem at hand,
but APGA is not. What we also show is that APGA (or any other GA) can be
faster than the parameter-less GA, but it needs to be tweaked to do that.

\section{Summary and Conclusions} \label{summaryAndConclusions}

This paper revisited two algorithms that resize the population during the
EA run itself. It was shown that one of these algorithms, the APGA, is not
capable of properly adapting the population size, and that its newly
introduced parameters act as the actual population size. This behavior is
independent of the problem being solved, is supported on theoretical grounds,
and confirmed by computer simulations.

This paper also raises important issues regarding fairness in empirical
comparative studies. The utilization of test problem generators eliminate
to some extent the degree of tweaking that can be done to make a particular
algorithm beat another algorithm. But we have shown that the utilization
of a test problem generator by itself is not sufficient to make fair
empirical comparisons.

The population plays a very important role in an evolutionary algorithm
and it is unfortunate that is continues to be largely underestimated and
poorly understood by many.

%
%
%

We could not disagree more with the conclusions drawn in previous research
studies~\cite{Back:00,Eiben:04}. As of yet, the lifetime
principle has not shown to be an effective method for adapting the
population size, and fixes such as those incorporated in APGA constitute
a poor implementation of that general idea.


\section*{Acknowledgments}
The authors thank the support of the Portuguese 
Foundation for Science and Technology (FCT/MCES) under grants 
POSI/SRI/42065/2001, POSC/EEA-ESE/61218/2004, and SFRH/BD/16980/2004.

\bibliographystyle{my-apa-uiuc}
\bibliography{references}

\end{spacing}
\end{document}